\documentclass{article}

\usepackage[preprint]{neurips_2024}

\usepackage[utf8]{inputenc} 
\usepackage{booktabs}       
\usepackage{amsfonts}       
\usepackage{nicefrac}       
\usepackage{microtype}      
\usepackage{xcolor}         
\usepackage{amsmath}
\usepackage[authoryear]{natbib}
\usepackage{graphicx}
\usepackage{wrapfig}
\usepackage{bm}
\usepackage{pifont}
\usepackage{cases}
\usepackage{algorithm}
\usepackage{enumitem}
\usepackage{multirow}

\title{Personalizing Low-Rank Bayesian Neural Networks \\ Via Federated Learning}

\author{%
  Boning Zhang \\
  University of Glasgow\\
  \texttt{b.zhang.6@research.gla.ac.uk} \\
  \And
  Dongzhu Liu \\
  University of Glasgow\\
  \texttt{dongzhu.liu@glasgow.ac.uk} \\
  \AND
  Osvaldo Simeone \\
  King's College London \\
  \texttt{osvaldo.simeone@kcl.ac.uk} \\
  \And
  Guanchu Wang \\
   Rice University\\
  \texttt{hegsns@rice.edu} \\
  \And
  Dimitrios Pezaros \\
  University of Glasgow \\
  \texttt{dimitrios.pezaros@glasgow.ac.uk} \\
    \And
  Guangxu Zhu \\
  Shenzhen Research Institute of Big Data \\
  \texttt{gxzhu@sribd.cn} \\
}

\begin{document}

\maketitle

\begin{abstract}
To support real-world decision-making, it is crucial for models to be well-calibrated, i.e., to assign reliable confidence estimates to their predictions. Uncertainty quantification is particularly important in personalized federated learning (PFL), as participating clients typically have small local datasets, making it difficult to unambiguously determine optimal model parameters. Bayesian PFL (BPFL) methods can potentially enhance calibration, but they often come with considerable computational and memory requirements due to the need to track the variances of all the individual model parameters. Furthermore, different clients may exhibit heterogeneous uncertainty levels owing to varying local dataset sizes and distributions. To address these challenges, we propose LR-BPFL, a novel BPFL method that learns a global deterministic model along with personalized low-rank Bayesian corrections. To tailor the local model to each client's inherent uncertainty level, LR-BPFL incorporates an adaptive rank selection mechanism. We evaluate LR-BPFL across a variety of datasets, demonstrating its advantages in terms of calibration, accuracy, as well as computational and memory requirements.
\end{abstract}

\section{Introduction} \label{sec:intro}
Federated Learning (FL) has emerged as a powerful distributed learning framework, enabling multiple clients to collaboratively train a global model without sharing their local datasets. FL has demonstrated success across various domains, including healthcare \citep{xu2021federated}, finance \citep{li2018measuring}, the Internet of Things \citep{nguyen2021federated}, and computer vision \citep{oh2021fedbabu}. However, when data distributions vary significantly across clients, the traditional FL framework, which focuses on training a single global model, often struggles to generalize well to individual clients, leading to performance degradation \citep{zhao2018federated, tan2022towards}. 

\begin{figure}
\centering
\includegraphics[width=0.5\linewidth]{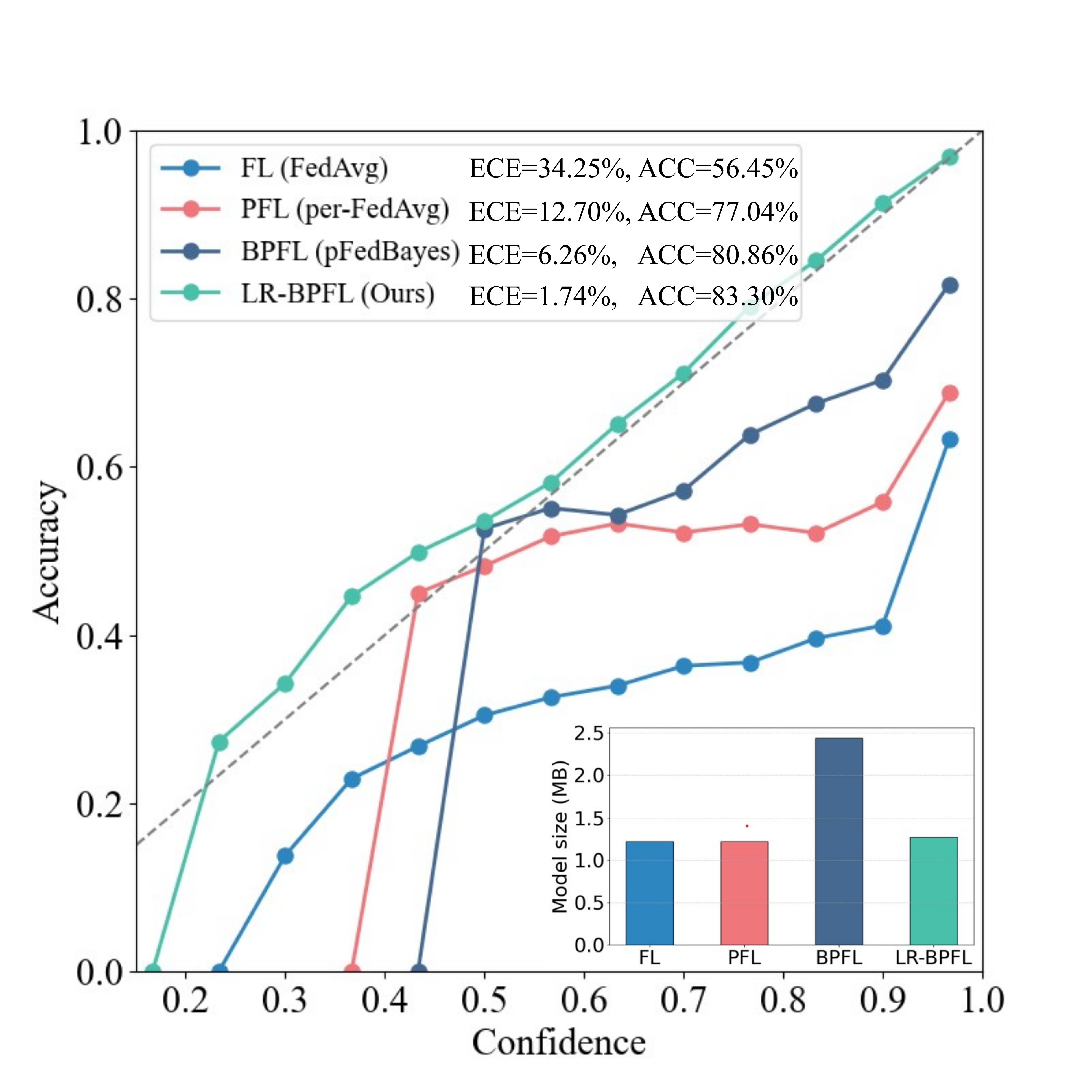}
\vspace{-5pt}
\caption{Average reliability diagrams across clients on the CIFAR-10 dataset for FedAvg \citep{mcmahan2017communication}, a standard FL algorithm; Per-FedAvg \citep{fallah2020personalized}, a benchmark PFL scheme; pFedBayes \citep{zhang2022personalized}, a state-of-the-art BPFL algorithm; and the proposed LR-BPFL. (See Section \ref{sec:uc} for details.)}
\vspace{-25pt}
\label{fig:intro_effect}
\end{figure}

Personalized Federated Learning (PFL) \citep{t2020personalized, li2020federated} addresses this issue by allowing each client to optimize a personalized model while still leveraging shared information collected from other clients through FL. Despite the benefits of personalization in accounting for data heterogeneity, PFL methods are typically trained on smaller effective datasets, which often exacerbates the uncertainty in the model-parameter space. Moreover, standard PFL methods frequently output models that fail to correctly quantify local uncertainty, producing overconfident decisions \citep{fallah2020personalized, collins2021exploiting}. This is illustrated in Fig. \ref{fig:intro_effect} for FedAvg \citep{mcmahan2017communication}, a standard FL method, and for Per-FedAvg \citep{fallah2020personalized}, a benchmark PFL scheme.

Bayesian learning can potentially produce well-calibrated models by representing uncertainty in the model-parameter space. Recent efforts to integrate Bayesian learning into PFL have shown promise in enhancing model performance, offering improved predictive accuracy and calibration in heterogeneous data scenarios \citep{zhang2022personalized, achituve2021personalized, jeon2024federated}. This is illustrated in Fig. \ref{fig:intro_effect} for pFedBayes \citep{zhang2022personalized}, a representative BPFL algorithm. However, these methods often rely on full-dimensional posterior distributions, leading to substantial computational and memory requirements. For example, in \cite{zhang2022personalized}, the distribution over the model parameters is specified by weight-specific means and variances, thus doubling the number of parameters that need to be stored and processed.

Rank-1 Bayesian Neural Networks (BNNs) \citep{dusenberry2020efficient} provide a parameter-efficient alternative to full BNNs by modeling the posterior distribution only over a rank-1 correction of deterministic model weights. However, incorporating rank-1 BNNs into FL is non-trivial, as it requires new strategies for model sharing and for personalization in order to address the heterogeneous distributions and sizes of the local datasets. 

Thus motivated, in this paper, we propose LR-BPFL, a novel BPFL scheme that learns a global deterministic model along with personalized low-rank Bayesian corrections. To tailor the local model to each client's inherent uncertainty level, LR-BPFL incorporates an adaptive rank selection mechanism. As exemplified in Fig. \ref{fig:intro_effect}, LR-BPFL is demonstrated via experiments to enhance accuracy and calibration as compared to the state-of-the-art methods, while also significantly reducing the model size, and, with it, computational and memory requirements. 

The main contributions are summarized as follows: 

\noindent $\bullet$ We introduce LR-BPFL, a BPFL protocol that reduces computational and memory requirements by jointly learning a shared deterministic model, while enabling clients to retain personalized Bayesian low-rank corrections. 

\noindent $\bullet$ To address data heterogeneity, LR-BPFL incorporates a novel adaptive rank selection mechanism, which dynamically adjusts the complexity of the Bayesian corrections based on client-specific uncertainty levels.

\noindent $\bullet$ We empirically evaluate LR-BPFL on several real-world datasets, demonstrating significant improvements in calibration and predictive accuracy as compared to state-of-the-art FL, BFL, PFL and BPFL methods, particularly in scenarios with small and heterogeneous client datasets (see Fig. \ref{fig:intro_effect}).

\section{Related Work}\label{sec:related_work}

\textbf{Personalized federated learning.}
PFL methods have been devised based on different principles, including local customization \citep{li2020federated, hanzely2020federated}, multi-task learning \citep{smith2017federated, marfoq2021federated}, meta-learning \citep{fallah2020personalized, chen2018federated}, client clustering \citep{briggs2020federated, sattler2020clustered}, and model mixing \citep{collins2021exploiting, deng2020adaptive}. While these methodologies enhance predictive accuracy on non-i.i.d. data, they offer suboptimal performance in terms of calibration \citep{achituve2021personalized}.

\textbf{Bayesian Neural Network.} BNNs combine the flexibility and expressive power of neural networks with the probabilistic properties of Bayesian inference, providing a powerful tool for uncertainty modeling \citep{neal2012bayesian}. By accounting for uncertainty in the model-parameter space, BNNs enhance calibration performance \citep{khan2021bayesian, knoblauch2019generalized, simeone2022machine}. However, the high computational and memory complexity of BNNs has hindered their widespread adoption, particularly for resource-constrained clients. To address this limitation, one line of research advocates the use of sparsity-promoting priors to reduce computational load by pruning neurons or weights  \citep{molchanov2017variational, bai2020efficient}. Another direction of research focuses on selectively applying Bayesian inference to critical parts of the model, while maintaining a deterministic, frequentist approach for other parts of the model  \citep{daxberger2021laplace}. A further promising strategy involves low-rank approximation. Notably, \cite{dusenberry2020efficient} proposes a rank-1 parameterization of BNNs.

\textbf{Bayesian federated learning.}
While traditional FL methods typically focus on point estimates of model parameters, recent progress has applied Bayesian learning to FL. Bayesian FL (BFL) aims to derive a global posterior distribution that aggregates knowledge from all participating clients within the federated network. Algorithms, like FedPA \citep{al2020federated}, QLSD \citep{vono2022qlsd} and DSVGD \citep{kassab2022federated}, employ Bayesian posterior decomposition techniques to break down the global posterior into the product of local posteriors; while FedBE \citep{chen2020fedbe} constructs a Gaussian or Dirichlet distribution at the server for Bayesian model ensembling. 

\textbf{Bayesian Personalized Federated Learning.} To address statistical heterogeneity among clients, pFedGP \citep{achituve2021personalized} learns a shared deep kernel function for all clients, while maintaining a personalized Gaussian process classifier at each client. Similarly, pFedBayes \citep{zhang2022personalized} enables each client to maintain its own personalized BNN, while using the aggregated global posterior as the prior distribution. Furthermore, MetaVD \citep{jeon2024federated} employs a hypernetwork to predict client-specific posterior distributions of Gaussian noises. 

\section{Preliminaries}\label{sec:preliminary}
This section reviews the standard FL setup \citep{mcmahan2017communication}, BPFL via variational inference \citep{zhang2022personalized, jeon2024federated}, as well as the rank-1 BNN method \citep{dusenberry2020efficient} as required preliminaries.

\subsection{Federated Learning}
The typical FL setup involves a server coordinating with $K$ clients to collaboratively train a global model without data sharing. Consider the training of a neural network parameterized by the weight vector $\mathbf{W}$. Each client $k$ has a loss function $F_k(\cdot)$ used for training on its local dataset $\mathcal{D}_k = \{({\bf x}_j, {\bf y}_j)\}_{j=1}^{|\mathcal{D}_k|}$.  In conventional FL, the objective for all clients is to find a global model that minimizes the weighted sum
\begin{equation}\label{equ:fedavg}
\underset{\mathbf{W}}{\operatorname{min}} \sum_{k=1}^K l_k F_k({\bf W}, \mathcal{D}_k), 
\end{equation}
where the weight $l_k$ is typically proportional to the size of the local dataset (i.e., $l_k=\left| \mathcal{D}_{k}\left|/ \right| \mathcal{D} \right|,$ with $\mathcal{D} = \{ \mathcal{D}_{k} \}_{k=1}^K$). 
To address this problem, in each communication round $t$, a subset $\mathcal{S}^t$ of clients is selected to conduct local training, starting from the latest global model weights ${\bf W}^{t}$.  At the end of the communication round $t$, the server aggregates the local models from the selected clients to update the global model, e.g., ${\bf W}^{t+1}  \leftarrow \sum_{k \in \mathcal{S}^t} l_k {\bf W}_{k}^{t}$, where ${\bf W}_k^{t}$ is the updated parameter vector produced by client $k$.

\subsection{Bayesian Personalized Federated Learning via Variational Inference}\label{sec:vi}
Variational inference methods approximate posterior distributions in the model-parameter space for a BNN, by using a family of variational distributions $q(\mathbf{W}; \bm{\phi})$, which depends on a set of variational parameters $\bm{\phi}$ (see, e.g., \citep{bishop2006pattern}; \citep{simeone2022machine}). For each client $k$, variational inference aims at minimizing the negative evidence lower bound (ELBO), also known as free energy, as in
\begin{equation}
\begin{aligned}\label{equ:vi_fl}
\min _{\bm{\phi}_k} \big\{- \mathbb{E}_{q\left({\bf W}_k; \bm{\phi}_k \right)} \left[\log p\left(\mathcal{D}_k \mid {\bf W}_k \right) \right] + \operatorname{KL}\left(q\left({\bf W}_k; \bm{\phi}_k \right) \| p\left({\bf W}_k\right) \right)\big\}, 
\end{aligned}
\end{equation}
where $- \log p\left(\mathcal{D}_k \mid {\bf W}_k \right)$ represents the cross-entropy loss accrued by the local model parameter ${\bf W}_k$ on the local dataset $\mathcal{D}_k$;  $\operatorname{KL}(q \| p)$ is the Kullback-Leibler divergence between distributions $p$ and $q$; $\bm{\phi}_k$ is a vector of local variational parameters; and $p({\bf W}_k)$ represents a prior distribution. The free energy (\ref{equ:vi_fl}) is thus a regularized version of the average cross-entropy loss (the first term in (\ref{equ:vi_fl})), with regularization dictated by the prior distribution $p({\bf W}_k)$ (the second term in (\ref{equ:vi_fl})).

pFedBayes \citep{zhang2022personalized} optimizes jointly the global prior distribution $p({\bf W})$ and local posteriors $q({\bf W}_k; \bm{\phi}_k)$ for all clients $k=1,\cdots, K$ by setting $p({\bf W}_k) = p({\bf W})$ in (\ref{equ:vi_fl}). Assuming  the prior $p({\bf W})$ and the local posterior $q({\bf W}_k; \bm{\phi}_k)$ to be Gaussian with diagonal covariance, the number of model parameters to be stored and processed is thus doubled as compared to non-Bayesian PFL methods.
 
MetaVD \citep{jeon2024federated} models the local posterior distribution $q({\bf W}_k; \bm{\phi}_k)$ as a Gaussian with a common mean ${\bf W}$ and a client-specific diagonal covariance, where the variance terms are produced by a shared hypernetwork that takes a client-specific learnable embedding as input. Like pFedBayes, MetaVD doubles the number of parameters that need to be stored and processed.

\subsection{Rank-1 Bayesian Neural Network} \label{sec:rbnn}
As discussed in Section \ref{sec:vi}, a BNN maintains a distribution $q({\bf W; \bm{\phi}})$ over the model parameters at the cost of increased computational and memory requirements. For single-model training, rank-1 BNNs present an efficient alternative in which the weight matrix $\mathbf{G} \in \mathbb{R}^{m \times n}$ for each layer is parameterized as $\mathbf{G} = \mathbf{W} \circ \mathbf{q} \mathbf{r}^{{\sf T}}$, where $\mathbf{W}$ represents a deterministic weight matrix, while $\mathbf{q}$ and $\mathbf{r}$ are $m$- and $n$-dimensional vectors, respectively, that are treated as random variables within a Bayesian framework. The symbol $\circ$ represents the element-wise product. 

Training is based on the minimization of the free energy loss, i.e., 
\begin{align}\label{equ:vi_bnn} 
\underset{{\bf W}, \bm{\phi}, \bm{\psi}}{\min} \Big\{& - \mathbb{E}_{q(\mathbf{q; \bm{\phi}}), q(\mathbf{r; \bm{\psi}})} \left[\log p(\mathcal{D} \mid {\bf W} \circ {\bf q} {\bf r}^{\sf T} )\right]  \\
& +  {\rm KL}\big( q(\mathbf{q; \bm{\phi}}) \parallel p(\mathbf{q}) \big) +  {\rm KL}\big( q(\mathbf{r; \bm{\psi}}) \parallel p(\mathbf{r}) \big) \Big\}, \nonumber 
\end{align}
where $\mathcal{D}$ is the dataset available at a given client; $q(\mathbf{q; \bm{\phi}})$ and $q(\mathbf{r; \bm{\psi}})$ are the variational distributions parameterized by vectors $\bm{\phi}$ and $\bm{\psi}$; and $p(\mathbf{q})$ and $p(\mathbf{r})$ are prior distributions. Rank-1 BNNs support the BatchEnsemble method \citep{wen2020batchensemble}, in which the output of multiple models sampled from the variational posteriors can be evaluated simultaneously for a given mini-batch, enabling efficient parallel computations.

\section{Low-rank Bayesian Personalized Federated Learning}\label{sec:lr-bpfl}
This section introduces LR-BPFL, a novel BPFL protocol that optimizes a shared model along with low-rank Bayesian multiplicative corrections for each client, whose rank is adapted to each client's local uncertainty levels. The overall protocol is detailed in Algorithm \ref{algo:LR-BPFL}.

\subsection{Low-Rank BNNs With Adaptive Rank Selection}
LR-BPFL describes the $m \times n$ weight matrix ${\bf G}_k$ for each layer of client $k$'s model as 
\begin{equation}\label{equ:lrbnn}
{\bf G}_k = {\bf W} \circ {\bf M}_k,
\end{equation} 
where ${\bf W}$ represents the shared deterministic model, while ${\bf M}_k$ denotes the personalized Bayesian correction. To support personalization of uncertainty quantification, we model the correction ${\bf M}_k$, also referred to as the Bayesian mask, as 
\begin{equation}\label{equ:mask}
{\bf M}_k = {\bf Q}_k {\bf \Lambda}_k {\bf R}_k^{\sf T},
\end{equation}
where ${\bf Q}_k$ is an $m \times r_{\max}$ matrix, ${\bf R}_k$ is a $n \times r_{\max}$ matrix, and ${\bf \Lambda}_k$ is an $r_{\max} \times r_{\max}$ gating matrix. The parameter $r_{\max}$ represents the maximum rank for the mask ${\bf M}_k$, and is set to be much smaller than $\min(m, n)$. To select the optimal rank, the diagonal gating matrix ${\bf \Lambda}_k$ has binary diagonal elements $\lambda_{k, i} \in \left\{0, 1 \right\}$ that control whether the corresponding $i$-th column of the dictionary matrices ${\bf Q}_k$ and ${\bf R}_k$ are retained (when $\lambda_{k, i}=1$) or pruned (when $\lambda_{k, i}=0$). The rank of the personalized Bayesian mask ${\bf M}_k$ is then given by 
\begin{equation}\label{equ:rank}
r_k = \sum_{i=1}^{r_{\max}} \lambda_{k, i}.
\end{equation}

\begin{algorithm}[t]
{\bf Input:} Number of global communication rounds $T$, number of clients $K$, fraction ratio $\tau$, learning rate $\eta$, size of the local dataset $N_k$, threshold $\bar{\lambda}$,  number of local adaptation steps $T_{L}$

{\bf Output:} ${\bf W}^{\star}, \left\{ {\bm \phi}_k^{\star}, {\bm \psi}_k^{\star}, {\bm \lambda}_k^{\star}\right\}_{k=1}^{K}$

{\bf Initialize:} Shared model $\mathbf{W}^{0}$, $\left\{ {\bm \phi}_k, {\bm \psi}_k, {\bm \lambda}_k\right\}_{k=1}^{K}$

{\bf for} round $t=0,\dots,T$ {\bf do}

\quad $\mathcal{S}^{t} \leftarrow$ Randomly sample $\tau K$ clients

\quad {\bf for} client $k \in \mathcal{S}^{t}$ {\bf in parallel do}

\quad\quad Receive $\mathbf{W}^{t}$ from the server and reinitialize ${\bm \phi}_k, {\bm \psi}_k$

\quad\quad $\left(\!{\bm \phi}_k,\! {\bm \psi}_k, \! {\bm \lambda}_k \! \right) \! \xleftarrow{T_L \text{steps}} \! \left(\!{\bm \phi}_k,\! {\bm \psi}_k, \! {\bm \lambda}_k \! \right) - \eta \nabla \! F_k\left({\bm \phi}_k, {\bm \psi}_k, {\bm \lambda}_k \right)$

\quad\quad Set $\mathbf{W}_k^{t} \leftarrow \mathbf{W}^{t} - \eta \nabla_{\mathbf{W}^{t}} F_{k}  \big({\bf W}^t, \bm{\phi}_k, \bm{\psi}_k, \bm{\lambda}_k \big)$

\quad\quad Update $\bm{\lambda}_k$ using (\ref{equ:thres})
 
\quad\quad Send $\mathbf{W}_{k}^{t}$ back to the server

\quad {\bf end for}

\quad Server aggregates local models as $\!\! \mathbf{W}^{t+1}\!\!\! = \!\! \sum_{k \in S^{t}} \! l_k \! \mathbf{W}_{k}^{t}$

{\bf end for}

\caption{LR-BPFL}\label{algo:LR-BPFL}

\end{algorithm}

\subsection{LR-BPFL}\label{sec:fl_lrbnn}
As detailed in Algorithm \ref{algo:LR-BPFL}, LR-BPFL learns a shared deterministic model ${\bf W}$ by following a standard frequentist approach, while the distributions $q({\bf Q}_k; \bm{\phi}_k)$ and $q({\bf R}_k; \bm{\psi}_k)$ of the Bayesian mask are optimized via variational learning through their variational parameters $\bm{\phi}_k$ and $\bm{\psi}_k$ (see Section \ref{sec:rbnn}). Accordingly, the learning problem is formulated as
\begin{align}
& \min _{{\bf W}, \{\bm{\phi}_k, \bm{\psi}_k, \bm{\lambda}_k\}_{k=1}^{K}}  \sum_{k=1}^K l_k F_{k}  \big({\bf W}, \bm{\phi}_k, \bm{\psi}_k, \bm{\lambda}_k \big),   \label{equ:objective} 
\end{align}
where $F_{k}(\cdot)$ denotes the regularized free energy loss in (\ref{equ:vi_bnn}), i.e., 
\begin{align}\label{equ:vi_loss}
F_{k} \big({\bf W}, \bm{\phi}_k, \bm{\psi}_k, \bm{\lambda}_k \big) =  &- \mathbb{E}_{q(\mathbf{Q}_k; \bm{\phi}_k), q(\mathbf{R}_k; \bm{\psi}_k)} \left[\log p(\mathcal{D}_k \mid {\bf W} \circ {\bf Q}_k {\bf \Lambda}_k {\bf R}_k^{\sf T} )\right] \\ 
& + {\rm KL}\big( q(\mathbf{Q}_k; \bm{\phi}_k) \parallel p(\mathbf{Q}_k) \big) + \nonumber
{\rm KL}\big( q(\mathbf{R}_k; \bm{\psi}_k) \parallel p(\mathbf{R}_k) \big),
\end{align}
and $\bm{\lambda}_k$ represents the diagonal vector of the matrix ${\bf \Lambda}_k$.

To solve this problem, we adopt a coordinate descent strategy, decoupling the updates of the Bayesian mask parameters $\{\bm{\phi}_k, \bm{\psi}_k, \bm{\lambda}_k \}_{k=1}^{K}$ from the shared model ${\bf W}$, along with a continuous relaxation of the binary variables $\{\bm{\lambda}_{k}\}_{k=1}^{K}$. The decoupling of updates allows the global model to capture general patterns across all clients via federated updates, while the Bayesian masks adapt to the local datasets.

Specifically, for a given shared model ${\bf W}$, each client $k$ addresses problem (\ref{equ:objective}) over the parameters $(\bm{\phi}_k, \bm{\psi}_k, \bm{\lambda}_k)$ by relaxing the entries of the vector $\bm{\lambda}_k$ as
\begin{equation}\label{equ:sigmoid}
\lambda_{k, i} = \sigma(\gamma_{k, i}),
\end{equation}
where $\sigma(\gamma) = (1 +\text{exp}(-\gamma))^{-1}$ is the sigmoid function, so that optimization is carried out over the corresponding unconstrained variables $\{\gamma_{k, i}\}_{i=1}^{r_{\max}}$. Furthermore, we model the variational distributions $q(\mathbf{Q}_k; \bm{\phi}_k), q(\mathbf{R}_k; \bm{\psi}_k)$ as Gaussian distributions with a trainable mean and diagonal covariance matrix. The priors $p({\bf Q}_k)$ and $p({\bf R}_k)$ also follow Gaussian distributions with fixed means and diagonal covariances. In a manner similar to \cite{hu2021lora}, we specifically set the means to $1/\sqrt{r_k}$, with $r_k$ being the current rank (\ref{equ:rank}). This way, the average value of the Bayesian mask elements is a priori equal to 1. The variance is set to a small number, such as 0.1.

Optimization over the parameters $( \bm{\phi}_k, \bm{\psi}_k) $ is performed by client $k$ using the reparameterization trick, and the KL divergence terms are evaluated in closed form (see, e.g., \cite{simeone2022machine}). To approximate the expected likelihood term in (\ref{equ:vi_loss}), we use Monte Carlo averaging, drawing $C$ samples from the variational distributions. To promote the selection of the most effective ranks, an L2-norm regularization term is applied to the vector $\bm{\lambda}_k$. 

After completing the local optimization of the parameters $(\bm{\phi}_k, \bm{\psi}_k, \bm{\lambda}_k)$, each active client updates the shared model ${\bf W}$ via stochastic gradient descent on the free energy (\ref{equ:vi_loss}). Subsequently, the rank $r_k$ is updated via a thresholding mechanism, with a fixed threshold $\bar{\lambda}$ as 
\begin{align}\label{equ:thres}
\lambda_{k, i} \leftarrow \lambda_{k, i} \cdot \mathbf{1}_{\{\lambda_{k, i} \geq \bar{\lambda}\}}, 
\end{align}
where $\mathbf{1}_{\{\cdot\}}$ is the indicator function and $\bar{\lambda}$ is a hyperparameter. We recommend setting the threshold $\bar{\lambda}$ close to 1 (e.g., 0.95).

Finally, the selected clients send their updated local models to the server, which then aggregates these models to obtain the next iterate.

\begin{figure*}[tbp]
\centering
\includegraphics[scale=0.2]{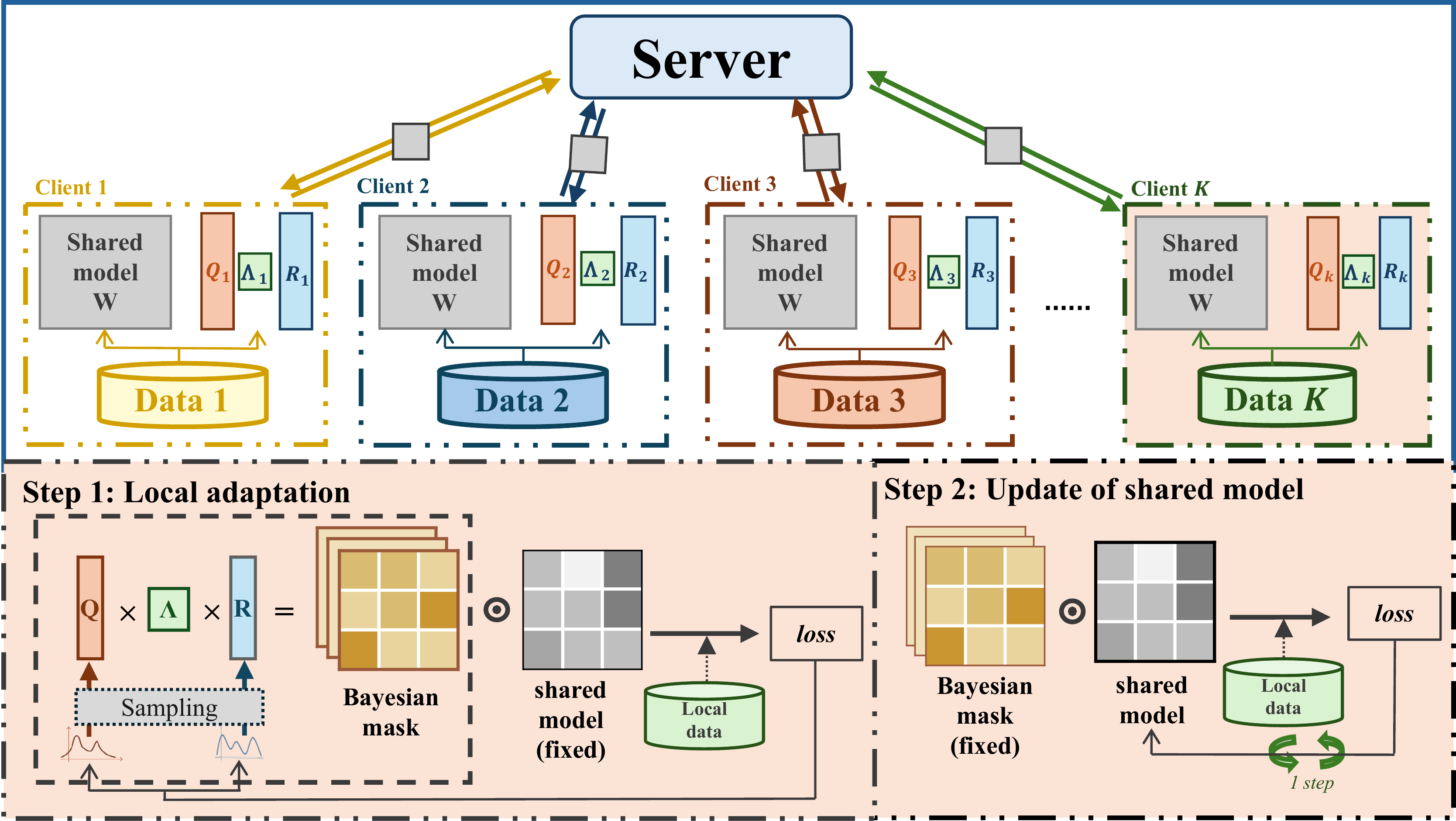}
\caption{\textbf{Top:} In the proposed LR-BPFL scheme, each client uploads its updated shared model to the server and then downloads the deterministic shared model from the server, while retaining and updating the Bayesian low-rank corrections locally. \textbf{Bottom:} During local training, after receiving the shared model from the server, clients perform local adaptation of the low-rank corrections before proceeding to update the shared model.}
\label{fig:system}
\end{figure*}

\section{Experiments}
To validate the effectiveness of LR-BPFL, this section reports on the results of extensive experiments across multiple scenarios, including non-i.i.d. data distributions, varying local dataset sizes, and different generalization requirements. Additionally, we performed an ablation study to assess the impact of the adaptive rank selection module on the model performance. We evaluated both accuracy and uncertainty calibration performance on several datasets, including CIFAR-10 and CIFAR-100.

\textbf{Baselines.} We compare LR-BPFL with the standard FL method FedAvg \citep{mcmahan2017communication}, the BFL algorithm FedBE \citep{chen2020fedbe}, PFL algorithms such as PerFedAvg \citep{fallah2020personalized} and pFedME \citep{t2020personalized}, as well as the BPFL algorithms MetaVD \citep{jeon2024federated}, pFedGP \citep{achituve2021personalized} and pFedBayes \citep{zhang2022personalized}.

\begin{table*}[t]
\caption{Comparison of computational, communication, memory requirements for different BPFL algorithms.}
\vspace{5pt}
  \label{results:overhead}
  \centering
  \begin{tabular}{cccc}
    \toprule
    Method & Training time (s)  & Parameters (M) & Communication (MB)\\
    \midrule
    MetaVD \citep{jeon2024federated} & 0.026 & 2.44 & 9.76  \\
    \midrule
    pFedGP \citep{achituve2021personalized} & 0.023 & {\bf 1.22} & {\bf 4.88}  \\
    \midrule
    pFedBayes \citep{zhang2022personalized} & 0.034 & 2.44  & 9.76 \\
    \midrule
    LR-BPFL (ours) & {\bf 0.018} & 1.27 & {\bf 4.88} \\
    \bottomrule
  \end{tabular}
\end{table*}

\begin{table*}
  \caption{Average Uncertainty calibration scores (ECE and MCE) evaluated on the CIFAR-10 and CIFAR-100 datasets. Lower scores indicate better calibration.} 
  \vspace{5pt}
  \label{result:cal}
  \centering
  \begin{tabular}{cccccccc}
    \toprule
    & Dataset & \multicolumn{4}{c}{CIFAR-10} & \multicolumn{2}{c}{CIFAR-100}\\
    \midrule
    & Data Heterogeneity & \multicolumn{2}{c}{2/10} & \multicolumn{2}{c}{5/10} & \multicolumn{2}{c}{5/100} \\
    \midrule
    & Method & ECE & MCE & ECE & MCE & ECE & MCE  \\
    \midrule
    FL & FedAvg \citep{mcmahan2017communication} & 0.343 & 0.495 & 0.382 & 0.517  & 0.623 & 0.745 \\
    \midrule
    BFL & FedBE \citep{chen2020fedbe} & 0.285 & 0.464 & 0.313 & 0.471 & 0.527 & 0.657 \\
    \midrule
    \multirow{2}{*}{PFL} & Per-FedAvg \citep{fallah2020personalized} & 0.127 & 0.222 & 0.316 & 0.438 & 0.353 & 0.491 \\
    & pFedMe \citep{t2020personalized} & 0.224 & 0.318 & 0.265 & 0.445 & 0.368 & 0.536 \\
    \midrule
    \multirow{4}{*}{BPFL} & MetaVD \citep{jeon2024federated} & 0.215 & 0.337 & 0.231 & 0.385 & 0.343 & 0.516 \\
    & pFedGP \citep{achituve2021personalized} & 0.169 & 0.315 & 0.210 & 0.333 & 0.265 & 0.446\\
    & pFedBayes \citep{zhang2022personalized} & 0.062 & 0.287 & 0.082 & 0.322 & 0.148 & 0.256\\
    & LR-BPFL (Ours) & {\bf 0.030} & {\bf 0.121} & {\bf 0.037} & {\bf 0.138} & {\bf 0.054} & {\bf 0.172} \\
    \bottomrule
  \end{tabular}
\end{table*}

\textbf{Experimental setting.} 
To maintain consistency with previous research, we employ a widely adopted CNN model across all algorithms and datasets. Specifically, the model comprises three convolutional layers, followed by three fully-connected layers for a total of 1.22 million parameters. For LR-BPFL, we set the maximum rank to $r_{\max}=8$, the variance of the priors $p({\bf Q}_k)$ and $p({\bf R}_k)$ to 0.1, the threshold for pruning ranks to $\bar{\lambda}=0.95$, and the ensemble size to $C=4$ for training and inference. Specifically, the predictive distribution is achieved by averaging the predictive probabilities from $C=4$ randomly generated Bayesian masks.

We adopt the non-i.i.d. setting from \cite{kotelevskii2022fedpop}. For CIFAR-10, we employ two configurations: in the first, each client has 2 out of the 10 labels, denoted as 2/10; and in the second, each client has 5 out of the 10 labels, denoted as 5/10. For CIFAR-100, each client has 5 out of the 100 labels, denoted as 5/100. Data heterogeneity is determined by the number of labels per client, and thus the 2-label setup is more non-i.i.d. than the 5-label setup. In all cases, labels are randomly selected from the label pools. Since small datasets pose a greater challenge to calibration, we train using only 10\% of the total training samples. In our experimental setup, we configure the number of communication rounds to $T=1000$, the number of clients to $K=50$, the fraction ratio of selected clients for each round to $\tau=0.2$, and the local update steps to 20. All experiments are conducted on a server equipped with an NVIDIA GeForce 4090 GPU, which includes 24GB of memory. For specific hyperparameter settings, please refer to the Appendix. 


\subsection{Computational, Memory, and Communication Requirements}
In this section, we compare the computational, memory, and communication requirements across the mentioned benchmark BPFL algorithms. Training time is measured as the average time per model update. As shown in Table \ref{results:overhead}, LR-BPFL achieves the shortest training time per step. Furthermore, in terms of memory requirements, LR-BPFL adds only 50K parameters — a 4\% overhead — significantly reducing memory usage as compared to other BFL algorithms. As for the communication overhead, since only the shared model is transmitted, the number of parameters sent remains identical to conventional FL algorithms. In contrast, both MetaVD and pFedBayes double the communication and memory overheads, due to the necessity of transmitting and storing both the means and variances for Bayesian updates. 

\begin{figure*}
\centering
\includegraphics[scale=0.11]{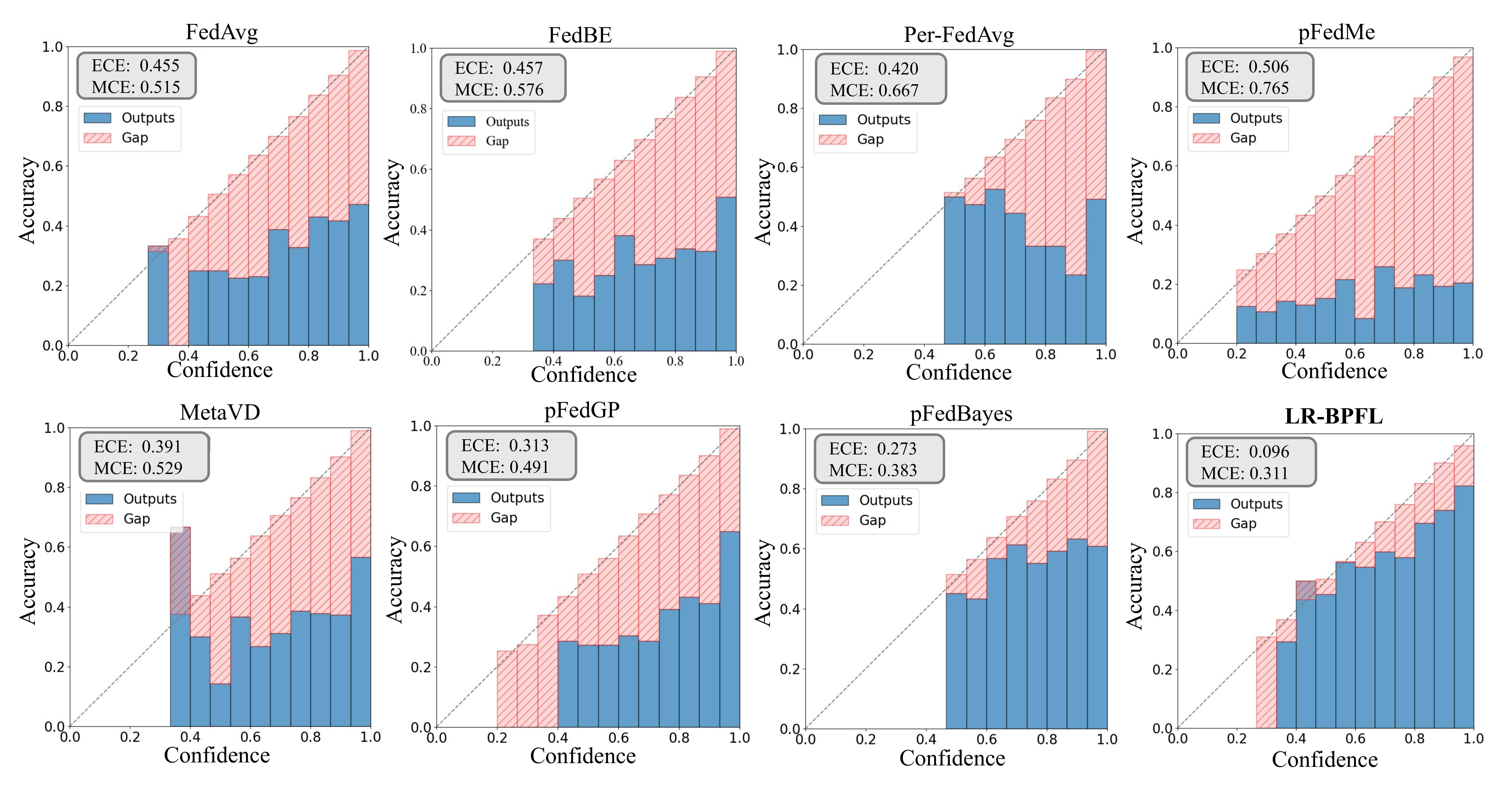}
\vspace{-10pt}
\caption{Reliability diagrams of the worst-calibrated client on CIFAR-10 with 50 clients. The diagonal line represents perfect calibration. Each plot also displays the ECE and MCE values.}
\label{fig:calibration}
\end{figure*}

\subsection{Uncertainty Calibration}\label{sec:uc}
Calibration is assessed using the expected calibration error (ECE) \citep{guo2017calibration}, which measures the average deviation between predicted probabilities and actual true class frequencies, and the maximum calibration error (MCE), which captures the largest deviation between predicted and actual classes. Table \ref{result:cal} summarizes the ECE and MCE results for the CIFAR-10 and CIFAR-100 datasets. The results indicate that BPFL algorithms generally achieve lower ECE and MCE values as compared to conventional FL algorithms such as FedAvg, PerFedAvg, and pFedMe.  LR-BPFL achieves the best ECE and MCE values among all the benchmark BPFL methods, highlighting its effectiveness in improving model calibration. 

To further investigate the impact of heterogeneity, Fig. \ref{fig:calibration} illustrates the worst calibration performance among all 50 participating clients. Compared to other methods, the reliability diagram for LR-BPFL remains closer to the diagonal line, indicating better calibration. This improvement in the worst client’s calibration performance underscores LR-BPFL's capability to achieve personalized calibration.

\subsection{Accuracy}
To evaluate the accuracy of the proposed method under non-i.i.d. data conditions, we conduct experiments using the same configuration described in the previous subsection, with datasets containing 40\% of the training samples. Table \ref{result:niid} presents the average classification accuracy on the CIFAR-10 and CIFAR-100 datasets across various non-i.i.d. settings defined by the ratio between labels present at a client and overall number of labels. As shown in Table \ref{result:niid}, PFL methods, such as pFedMe and pFedGP, generally outperform non-personalized FL methods like FedAvg. In all tested settings, LR-BPFL consistently outperforms other benchmark algorithms. 

\begin{table*}[t]
  \caption{Classification accuracies (in \%) with different heterogeneity degrees. Averages over 5 seeds are reported.}
  \vspace{5pt}
  \label{result:niid}
  \centering
  \begin{tabular}{c c c c }
    \toprule
    Dataset & \multicolumn{2}{c}{CIFAR10} & CIFAR100 \\
    \midrule
     Data heterogeneity  & 2/10 & 5/10 & 5/100  \\
    \midrule
    FedAvg \citep{mcmahan2017communication}  & 66.37 & 68.66 & 33.05  \\
    FedBE \citep{chen2020fedbe}   & 67.05 & 68.75 & 34.11  \\
    \midrule    
    Per-FedAvg \citep{fallah2020personalized}  & 82.38 & 68.91 & 60.24  \\
    pFedMe \citep{t2020personalized}  & 78.14 & 70.45 & 47.15  \\
    \midrule
    MetaVD \citep{jeon2024federated} & 67.52 & 68.58 & 34.23  \\
    pFedGP \citep{achituve2021personalized}   & 67.11 & 68.25 & 33.29  \\
    pFedBayes \citep{zhang2022personalized}   & 82.28 & 69.43 & 62.47   \\
    \midrule
    LR-BPFL (Ours)  & {\bf 84.67} & {\bf 72.38} & {\bf 66.12}  \\
    \bottomrule
  \end{tabular}
\end{table*}

\begin{table*}[h]
  \caption{Performance comparison of LR-BPFL with and without adaptive rank selection (LR-BPFL w/o ARS) on the CIFAR-10 and CIFAR-100 datasets. The table shows accuracy (in \%), average expected calibration error (A-ECE), and worst-client calibration error (W-ECE).}
  \vspace{5pt}
  \label{results:ablation}
  \centering
  \scalebox{0.8}{
  \begin{tabular}{cccccccccc}
    \toprule
     & \multicolumn{6}{c}{\emph{CIFAR-10} dataset}  & \multicolumn{3}{c}{\emph{CIFAR-100} dataset}\\
    \midrule
    & \multicolumn{3}{c}{2/10}  & \multicolumn{3}{c}{5/10} & \multicolumn{3}{c}{5/100}\\
    \midrule 
    Method & ACC  & A-ECE & W-ECE & ACC & A-ECE & W-ECE & ACC & A-ECE & W-ECE\\
    \midrule
    LR-BPFL (w/o ARS) & 82.62 & 0.035 & 0.130 & 64.88 & 0.044 & 0.143 & 55.64 & 0.061 & 0.156\\
    \midrule
    LR-BPFL & {\bf 83.30} & {\bf 0.030} & {\bf 0.096} & {\bf 65.68} & {\bf 0.038} & {\bf 0.111} & {\bf 56.49} & {\bf 0.054} & {\bf 0.124} \\
    \bottomrule
  \end{tabular}
  }
  \vspace{-5pt}
\end{table*}

\begin{figure}[t]
\vspace{-10pt}
\centering
\includegraphics[width = 0.7\linewidth]{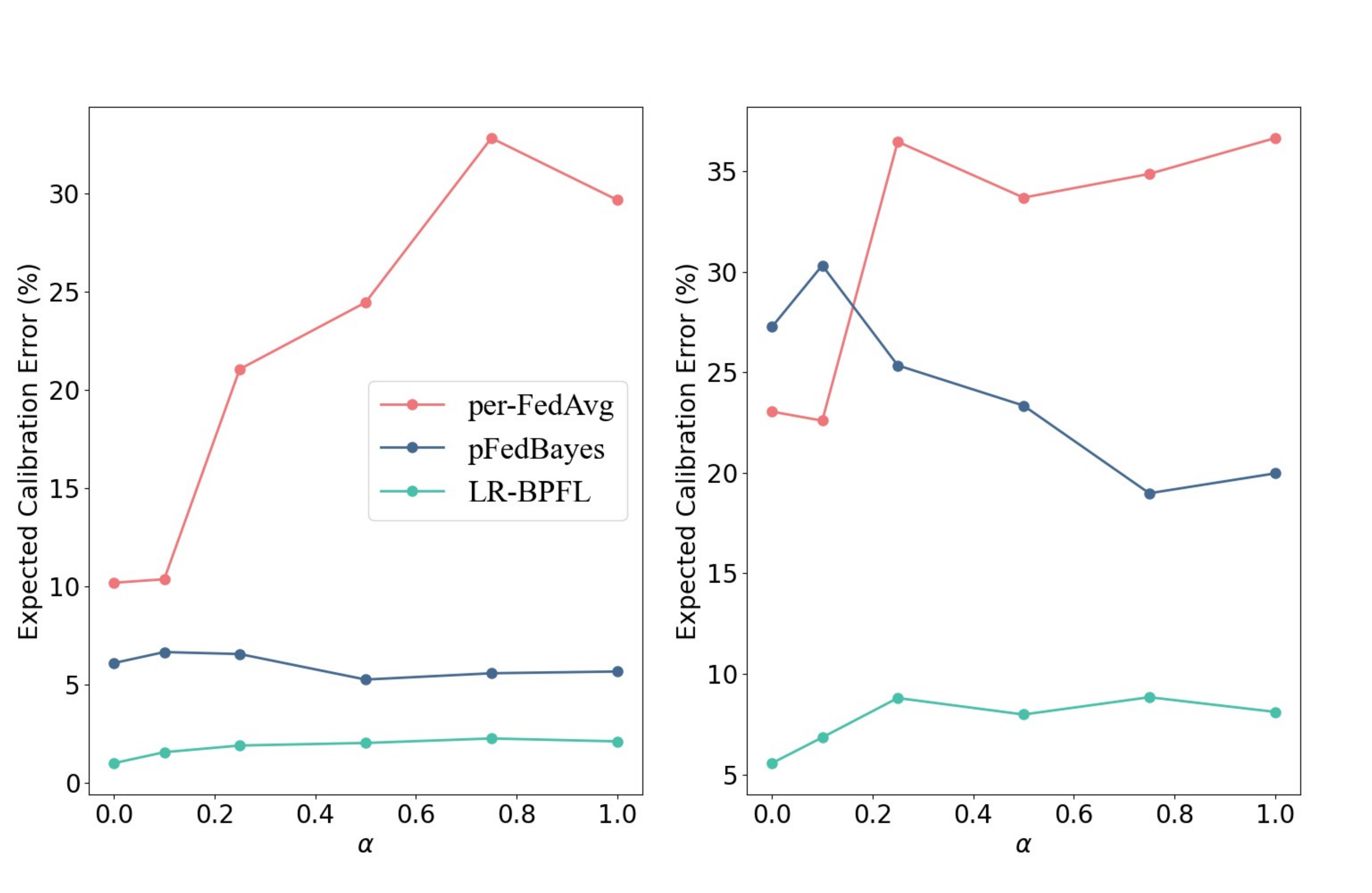}
\vspace{-15pt}
\caption{Calibration performance for new clients as a function of the parameter $\alpha$, which dictates the degree to which the local data distributions of the test clients differ from those of the training clients. As $\alpha$ moves away from 0.1, the distributions of new clients diverge more from those of the training clients.}
\label{fig:generalization}
\end{figure}
\subsection{Generalization to New Clients}
To evaluate calibration performance on unseen clients, which do not participate in training, we follow the evaluation method outlined in \cite{shamsian2021personalized}. Accordingly, the CIFAR-10 dataset is split into two distinct subsets: the first subset is distributed across 50 clients for model training, while the second is assigned to 10 additional clients for evaluation as new clients. Within both subsets, we set the class probabilities in each client by sampling from a Dirichlet distribution with the same $\alpha$ parameter. For the training subset, we set the parameter $\alpha=0.1$. During testing, we evaluate model performance on the new clients by varying  $\alpha \in \{0.1, 0.25, 0.5, 0.75, 1.0\}$.  As $\alpha$ moves away from 0.1, the distributions of new clients diverge more from those of the training clients, making adaptation to new clients more challenging. Fig. \ref{fig:generalization} illustrates the average ECE and worst-client ECE as a function of the Dirichlet parameter $\alpha$, where $\alpha=0$ corresponds to results from the clients involved in the training process. LR-BPFL consistently demonstrates superior calibration performance across all $\alpha$ values. Notably, even for the worst-performing clients, LR-BPFL maintains stable and reliable calibration. These findings highlight the robustness of our method in adapting to new clients while preserving calibration quality across all participants.

\begin{figure}[t]
\centering
\includegraphics[width=0.75\linewidth]{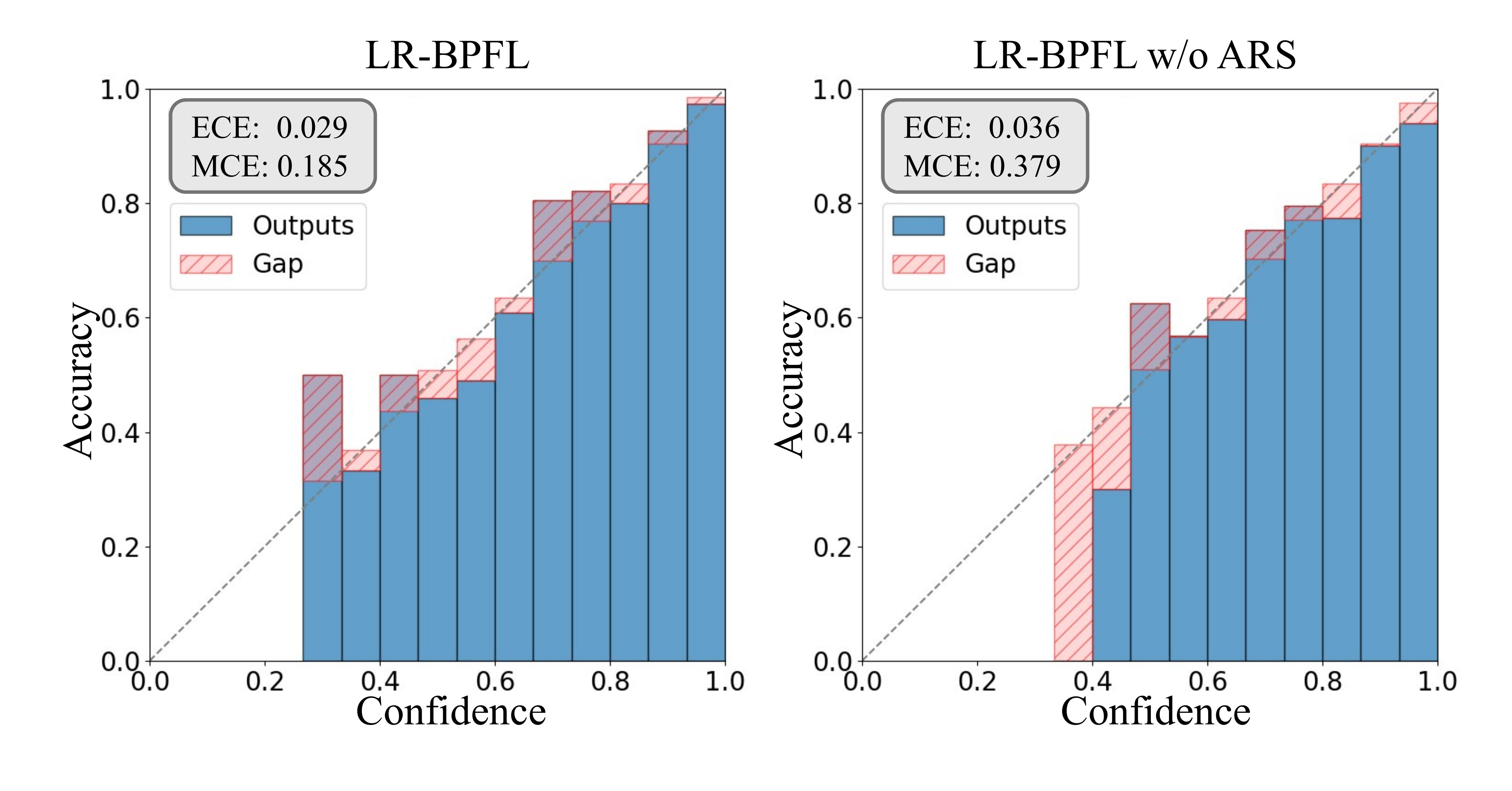}
\vspace{-5pt}
\caption{Comparison of reliability diagrams for LR-BPFL with and without the ARS module.}
\label{fig:ablation}
\end{figure}
\subsection{On the Role of Adaptive Rank Selection} \label{sec:ablation}
In this section, we conduct an ablation study to assess the effect of the adaptive rank selection (ARS) module in LR-BPFL. Table \ref{results:ablation} presents the classification accuracy, average ECE, and worst-client ECE for LR-BPFL with and without the ARS module. The results clearly demonstrate that incorporating the ARS module significantly enhances both accuracy and model calibration. Fig. \ref{fig:ablation} provides a visual comparison for a specific client, with the left figure representing results with ARS and the right figure representing results without it. The comparison highlights the significant contribution of ARS to overall performance.

\section{Conclusion}
We have proposed a novel personalized FL protocol that jointly learns a global deterministic model, along with personalized low-rank Bayesian corrections for each client. The local adaptation of Bayesian masks incorporates an adaptive rank selection module, which dynamically tailors the local model by adjusting the rank of the local masks. Our experiments validated that LR-BPFL can effectively address data heterogeneity, local epistemic uncertainty, and local computational and memory constraints, showcasing improved accuracy and calibration, as well as the capability to generalize to new clients as compared to the state of the art. Future work may explore fully decentralized versions of LR-BPFL.

\bibliography{refs}
\bibliographystyle{plainnat}

\newpage
\appendix
\section{Implementation Details}
\subsection{Baselines}
We present an overview of the baseline models used in our experiments, covering state-of-the-art approaches across four categories: FL, PFL, BFL, and BPFL methods.

The following \textbf{FL method} is considered in our experiments:
\begin{itemize}[leftmargin=1.2cm]
    \item FedAvg \citep{mcmahan2017communication} is a standard FL algorithm that averages gradients weighted by the data size of clients in each FL round.
\end{itemize}

The following \textbf{BFL method} is considered in our experiments:
\begin{itemize}[leftmargin=1.2cm]
    \item FedBE \citep{chen2020fedbe} enhances robust aggregation by sampling high-quality global models and combining them through Bayesian model ensembling, where the global model distribution is derived by considering each client's local model as a potential global model.
\end{itemize}

The following \textbf{PFL methods} are considered in our experiments:
\begin{itemize}[leftmargin=1.2cm]
    \item Per-FedAvg \citep{fallah2020personalized} applies model-agnostic meta-learning (MAML) to fine-tune the global model with local gradient updates, enabling personalized adaptation in federated learning.
    \item pFedMe \citep{t2020personalized} addresses personalized FL as a bi-level optimization problem, decoupling the personalized model optimization from global model learning by employing Moreau envelopes as the regularized loss function for each client.
\end{itemize}

The following \textbf{BPFL methods} are considered in our experiments:
\begin{itemize}[leftmargin=1.2cm]
    \item pFedGP \citep{achituve2021personalized} employs Gaussian process for personalized federated learning, leveraging shared kernel functions parameterized by a neural network and personalized Gaussian process classifier.  
    \item MetaVD \citep{jeon2024federated} models the local posterior distribution as a Gaussian, where the mean is derived from the deterministic global model, and the client-specific diagonal covariance is produced by a shared hypernetwork.
    \item pFedBayes \citep{zhang2022personalized} enables each client to maintain its own personalized BNN while using the aggregated global posterior as the prior distribution.
\end{itemize}

\subsection{Experimental hyperparameters}
For all datasets, we set $T = 1000$ to ensure convergence following standard conventions. The batch size is set to 32, and local steps are set to 20. In order to ensure a fair comparison between the algorithms, the results presented in all of our experiments are obtained using optimal hyperparameters for each model. 

For hyperparameter settings, we refer to the configurations in the corresponding papers of each algorithm and fine-tune the parameters around their recommended values. For instance, if an algorithm uses a learning rate $\eta = 0.01$, then we set the learning rate within the range of 0.001 to 0.1 for parameter tuning. For algorithm-specific hyperparameters, we use the values recommended in their papers.

Based on our experimental results, we set the learning rates for FedAvg to 0.1, and for Per-FedAvg and FedBE to 0.01. For pFedMe, the personalized learning rate, global learning rate, and regularization weight are set to 0.01, 0.01, and 15, respectively. The learning rate for pFedGP is set to 0.05. For pFedBayes, the tradeoff parameter is set to $\zeta = 10$ (see Eq. (27) in \cite{zhang2022personalized}), the learning rates for the personalized and global models are set to $\eta_1 = \eta_2 = 0.001$. For LR-BPFL, we set the learning rate for the global deterministic model to 0.001 and for the Bayesian mask to 0.01. To ensure reproducibility of the experiments, we will release all code on our GitHub repository.

\subsection{Dataset}
We provide a description of the datasets used in our experiments. The CIFAR-10 and CIFAR-100 datasets are commonly used for 10-class and 100-class image classification, respectively. Each contains 50,000 training images and 10,000 test images, all with a resolution of 32x32 pixels. In our experiments, we combine the training and test images, then distribute them to clients according to different non-i.i.d. settings to verify the impact of data heterogeneity. Finally, each client splits their assigned data into training and testing sets according to the specified percentage.

\section{Additional Experiments}
\subsection{Impact of local dataset sizes}
To evaluate the accuracy of the proposed method with varying local dataset sizes, we consider two configurations: the small dataset uses 10\% of the data for training, while the medium dataset uses 40\% of the data for training. The results are presented in Table \ref{result:data_size}. In the small dataset scenario, all schemes perform better with the more non-i.i.d. data distribution ( i.e., heterogeneity degree 2/10). A possible explanation is that when the dataset size is limited, a more i.i.d. distribution results in fewer samples per label for each client, which can lead to severe overfitting and, consequently, a decrease in accuracy. In all tested settings, our approach consistently outperforms other SOTA algorithms.

\begin{table}[h]
  \caption{Classification accuracies with different heterogeneity degrees and varying dataset sizes. The higher score, the better. Averages over 5 seeds are reported.}
  \label{result:data_size}
  \centering
  \begin{tabular}{c c c c c c c c c c}
    \toprule
    & \multicolumn{3}{c}{Small (Acc. (\%))} & \multicolumn{3}{c}{Medium (Acc. (\%))} \\
    \midrule
    Dataset & \multicolumn{2}{c}{CIFAR10} & CIFAR100 & \multicolumn{2}{c}{CIFAR10}& CIFAR100\\
    \midrule
     Data heterogeneity & 2/10 & 5/10 & 5/100 & 2/10 & 5/10 & 5/100  \\
    \midrule
    FedAvg \citep{mcmahan2017communication}  & 56.45 & 53.80 & 19.39 & 66.37 & 68.66 & 33.05  \\
    FedBE \citep{chen2020fedbe}  & 57.19 & 54.42 & 19.89 & 67.05 & 68.75 & 34.11  \\
    \midrule    
    PerFedAvg \citep{fallah2020personalized} & 77.04 & 54.84 & 45.49 & 82.38 & 68.91 & 60.24  \\
    pFedMe \citep{t2020personalized} & 61.37 & 57.53 & 37.27 & 78.14 & 70.45 & 47.15  \\
    \midrule
    MetaVD \citep{jeon2024federated}  & 57.02 & 54.58 & 20.74 & 67.52 & 68.58 & 34.23  \\
    pFedGP \citep{achituve2021personalized} & 55.14 & 52.27 & 19.05 & 67.11 & 68.25 & 33.29  \\
    pFedBayes \citep{zhang2022personalized} & 80.86 & 62.25 & 52.11 & 82.28 & 69.43 & 62.47   \\
    \midrule
    Ours   & {\bf 83.30} & {\bf 64.61} & {\bf 54.83} & {\bf 84.67} & {\bf 72.38} & {\bf 66.12}  \\
    \bottomrule
  \end{tabular}
\end{table}

\begin{figure}[h]
\centering
\includegraphics[width =0.55\linewidth]{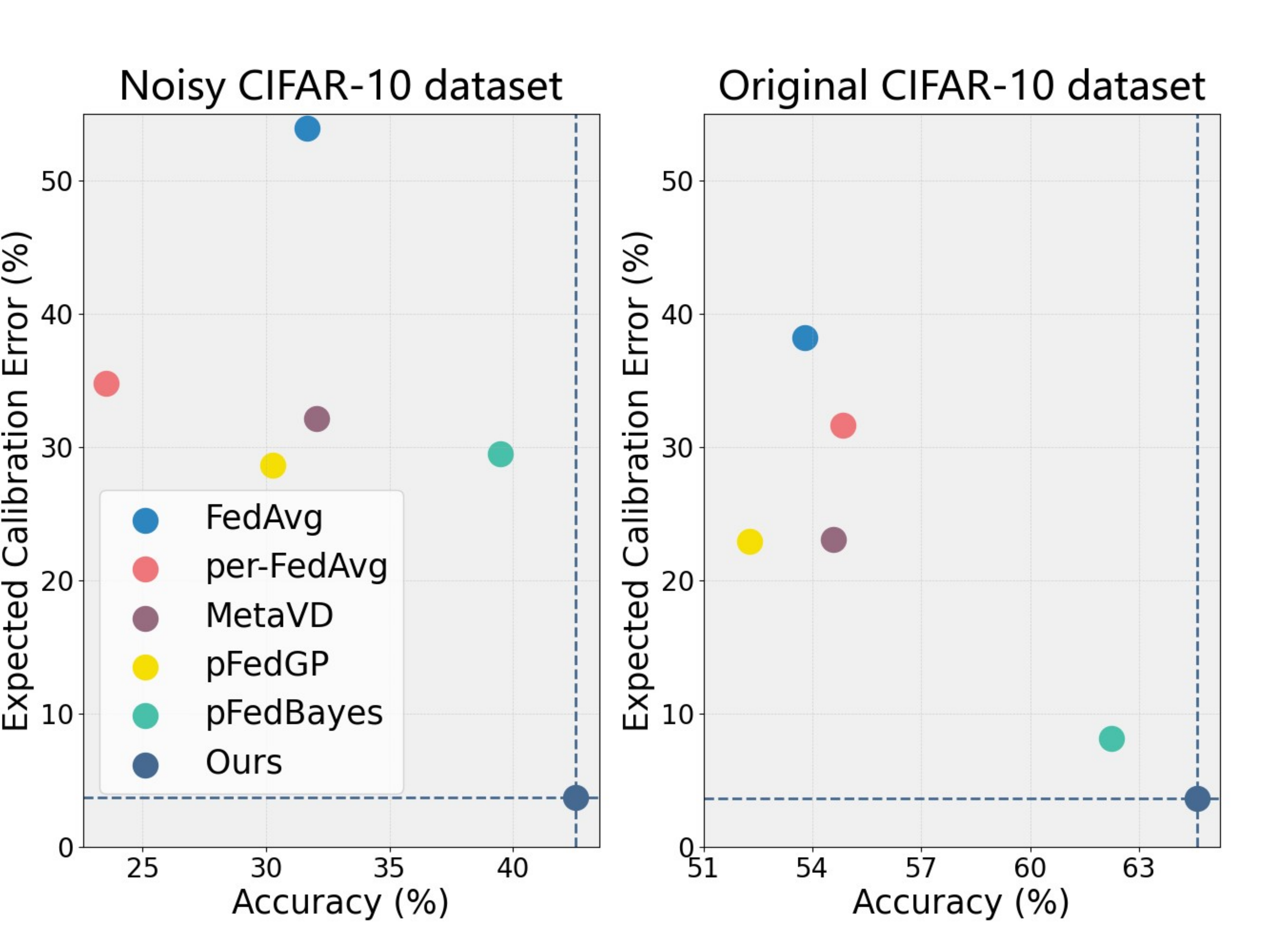}
\caption{Test accuracy and ECE (\%) on noisy and original CIFAR-10 dataset.}
\label{fig:noisy}
\end{figure}
\subsection{Impact of Noisy Dataset}
To investigate the impact of noisy data on our algorithm,  we use 10\% of the CIFAR-10 samples, with each client assigned 5 out of 10 labels, and apply 57 unique image corruption distributions as in \cite{achituve2021personalized}. Each client is assigned a distinct noise model, and within the client, each noisy data sample is randomly drawn based on the assigned model. The results, as illustrated in Fig. \ref{fig:noisy}, demonstrate the impact of noise on both accuracy and calibration. The accuracy is represented along the x-axis, while ECE is plotted on the y-axis, with better performance appearing closer to the lower right corner. The results show that models trained on the noisy dataset suffer significant degradation in accuracy compared to those trained on the original dataset. However, our approach calibrates the model well even under noisy datasets, remaining robust for providing reliable predictions despite showing a lower accuracy.

\subsection{On the Role of Coordinate Descent}
In this section, we investigate different update strategies by comparing the coordinate descent approach used in LR-BPFL with a joint update strategy, where the global deterministic model and the local Bayesian mask are updated jointly, but only the deterministic model is shared with the server during each communication round. This setup is referred to as LR-BPFL (joint). As shown in Table \ref{results:coordinate}, the coordinate descent approach leads to significant improvements in both accuracy and calibration. These results indicate that decoupling the updates of the global model and personalized masks allows the global model to focus on learning general patterns across all clients via federated updates, while the personalized Bayesian masks adapt to local datasets, resulting in enhanced overall performance.
\begin{table}[h]
  \caption{Performance comparison of LR-BPFL and LR-BPFL (joint) on the CIFAR-10 and CIFAR-100 datasets. The table shows accuracy (in \%), average expected calibration error (A-ECE), and worst-client calibration error (W-ECE).}
  \label{results:coordinate}
  \centering
  \small{
  \begin{tabular}{cccccccccc}
    \toprule
     & \multicolumn{6}{c}{\emph{CIFAR-10} dataset}  & \multicolumn{3}{c}{\emph{CIFAR-100} dataset}\\
    \midrule
    & \multicolumn{3}{c}{2/10}  & \multicolumn{3}{c}{5/10} & \multicolumn{3}{c}{5/100}\\
    \midrule 
    Method & ACC  & A-ECE & W-ECE & ACC & A-ECE & W-ECE & ACC & A-ECE & W-ECE\\
    \midrule
    LR-BPFL (joint) & 80.79 & 0.065 & 0.194 & 61.76 & 0.154 & 0.297 & 52.93 & 0.115 & 0.246\\
    \midrule
    LR-BPFL & {\bf 83.30} & {\bf 0.030} & {\bf 0.096} & {\bf 65.68} & {\bf 0.038} & {\bf 0.111} & {\bf 56.49} & {\bf 0.054} & {\bf 0.124} \\
    \bottomrule
  \end{tabular}
  }
\end{table}

\begin{figure}[t]
\centering
\includegraphics[width =0.6\linewidth]{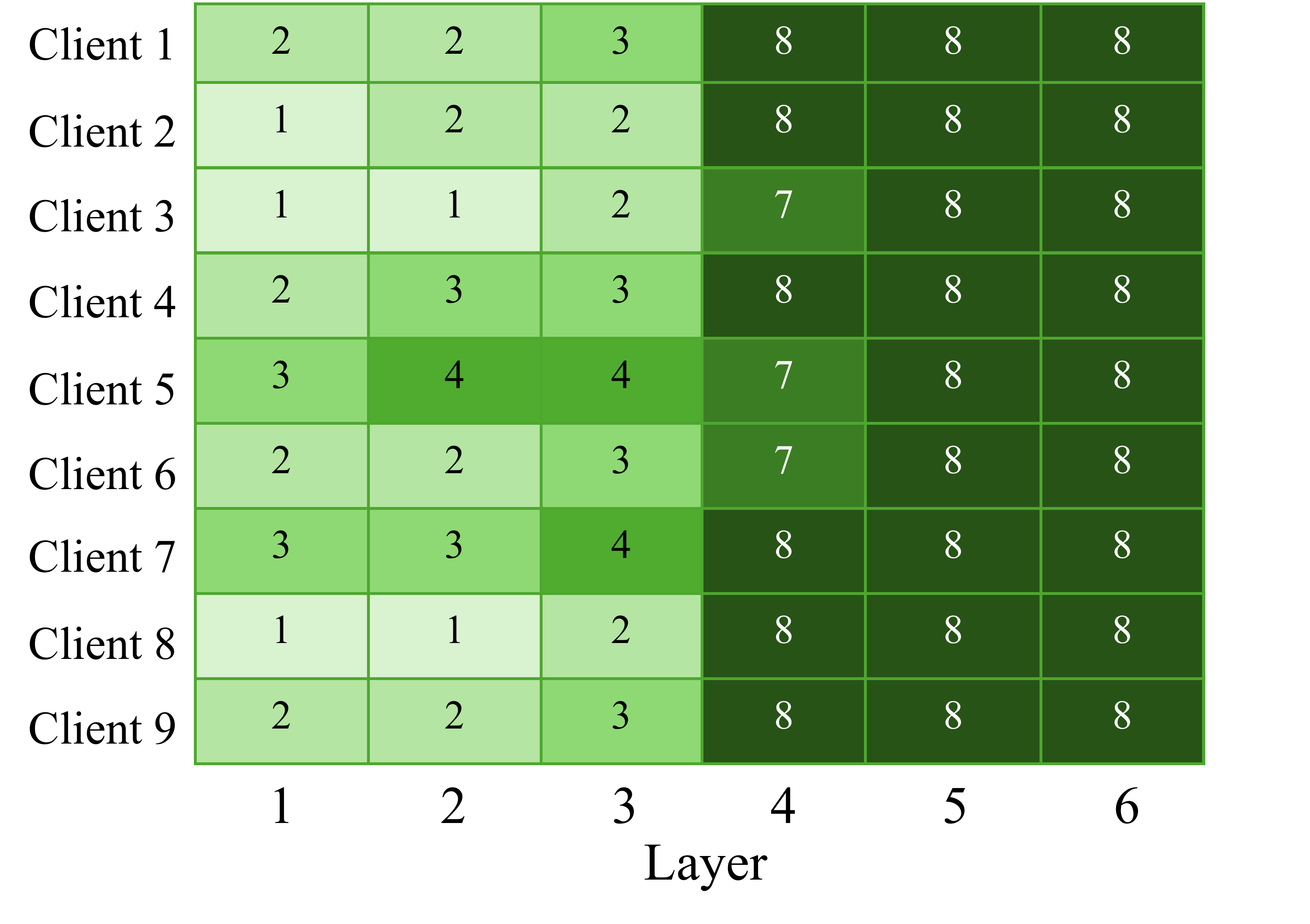}
\vspace{-10pt}
\caption{Resulting rank of each Bayesian mask based on the CIFAR-10 dataset using LR-BPFL. The x-axis shows the layer index, and the y-axis represents different clients.}
\label{fig:rank}
\end{figure}
\subsection{The effect of rank selection}
Since each Bayesian mask must retain at least one rank for personalization, we exclude the first rank from the pruning strategy. In our experiments, each binary element of the diagonal vector $\bm{\lambda}$ in the gating matrix is relaxed to $\lambda = \sigma(\gamma)$, where $\sigma$ is the sigmoid function. The initial value of $\gamma$ is set at 3.5, and we set the pruning threshold to 0.95. To encourage the selection of effective ranks, we apply an L2-norm regularization term with a weight of 0.1 to the vector $\bm{\lambda}$ in the loss function. We also set the learning rate to 0.001 for updating $\gamma$. Fig. \ref{fig:rank} illustrates an example of the final rank distribution per client. As shown, the optimal ranks vary across different layers and clients. This supports our hypothesis that clients require different BNN architectures to adapt to their uncertainty levels.

\end{document}